\documentclass[letterpaper, 10 pt, conference]{ieeeconf}  

\IEEEoverridecommandlockouts                              

\usepackage{graphics} 
\usepackage{epsfig} 
\usepackage{times} 
\usepackage{amsmath} 
\usepackage{amssymb}  
\usepackage{bm}      
\usepackage{multirow}
\usepackage{multicol}
\usepackage{makecell}
\usepackage{mathtools}
\usepackage{cite}

\usepackage{diagbox}

\usepackage{float}
\usepackage[caption=false]{subfig}
\usepackage{hyperref}
\usepackage{arydshln} 
\let\labelindent\relax
\usepackage{enumitem}

\usepackage[dvipsnames]{xcolor}
\definecolor{revised}{RGB}{0, 0, 0} 

\title{\LARGE \bf
Gradient Field-Based Dynamic Window Approach for Collision Avoidance in Complex Environments 
}

\author{Ze Zhang$^{1}$, Yifan Xue$^{2}$, Nadia Figueroa$^{2}$, and Knut Åkesson$^{1}$
\thanks{*This paper has been accepted by IEEE/RSJ International Conference on Intelligent Robots and Systems (IROS) 2025, Hangzhou.} 
\thanks{$^{1}$Ze Zhang and Knut Åkesson are with Chalmers University of Technology, 41296 Gothenburg, Sweden {\tt\scriptsize \{zhze, knut\}@chalmers.se}}%
\thanks{$^{2}$Yifan Xue and Nadia Figueroa are with the University of Pennsylvania, Philadelphia, PA
19104 USA {\tt\scriptsize\{yifanxue, nadiafig\}@seas.upenn.edu}}%
}

\begin{document}

\maketitle
\thispagestyle{empty}
\pagestyle{empty}

\begin{abstract}
For safe and flexible navigation in multi-robot systems, this paper presents an enhanced and predictive sampling-based trajectory planning approach in complex environments, the Gradient Field-based Dynamic Window Approach (GF-DWA). 
Building upon the dynamic window approach, the proposed method utilizes gradient information of obstacle distances as a new cost term to anticipate potential collisions. This enhancement enables the robot to improve awareness of obstacles, including those with non-convex shapes. 
The gradient field is derived from the Gaussian process distance field, which generates both the distance field and gradient field by leveraging Gaussian process regression to model the spatial structure of the environment.
Through several obstacle avoidance and fleet collision avoidance scenarios, the proposed GF-DWA is shown to outperform other popular trajectory planning and control methods in terms of safety and flexibility, especially in complex environments with non-convex obstacles. 
\end{abstract}

\section{INTRODUCTION}
With the rise of autonomous mobile robots, there is a growing demand for advanced and flexible collision-free navigation algorithms that allow robots to operate autonomously in complex environments.
In the context of local motion planning and control \cite{siegwart_2011_introduction}, existing approaches can be broadly classified into optimization-based, sampling-based, and learning-based methods. Optimization-based methods, such as Model Predictive Control (MPC) \cite{sathya_2018_embedded, Ze_2023_mpcwta} and control barrier functions \cite{Thirugnanam_2022_polycbf, xue_2025_mcbf}, generate optimal control inputs by formulating an optimization problem with a defined cost function and constraints. These methods are widely used due to their ability to ensure safety and stability in navigation tasks. Sampling-based methods, including the Dynamic Window Approach (DWA) \cite{fox_1997_dwa, Lee_2021_dwadyn, missura_2019_preddwa, yasuda_2023_stochdwa, dobrevski_2020_adaptivedwa} with consideration of kinematic constraint and rapidly-exploring randomized trees \cite{Perez_2012_lqrrrt, orthey_2023_samplingplan}, generate a set of control or state samples and select the most suitable one based on objective function evaluation and feasibility checks. Recently, learning-based methods, such as Deep Reinforcement Learning (DRL) \cite{Ceder_2024_ddpg, kaufmann_2023_drldrone, chen_2017_decentralizeddrl}, have been explored for motion planning and control, which can learn the optimal policy from the interaction with the environment.
While convex obstacles can be handled efficiently by most navigation algorithms, non-convex obstacles present significant challenges, particularly in local motion planning and control. When multiple robots operate in close proximity, non-convex obstacles may frequently emerge due to the spatial configuration of robots and static obstacles \cite{Ceder_2024_ddpg, xue_2025_mcbf}.
These obstacles are particularly difficult to avoid as they introduce local minima in the objective function, making it challenging for both optimization-based and sampling-based methods to find feasible solutions. Additionally, learning-based methods face difficulties in overcoming non-convex obstacles, as successful navigation in such environments requires an effective exploration strategy during training that incorporates sufficient randomness and a long-term reward structure to ensure adaptive decision-making.

\begin{figure}[t]
    \centering
    \includegraphics[width=0.99\linewidth]{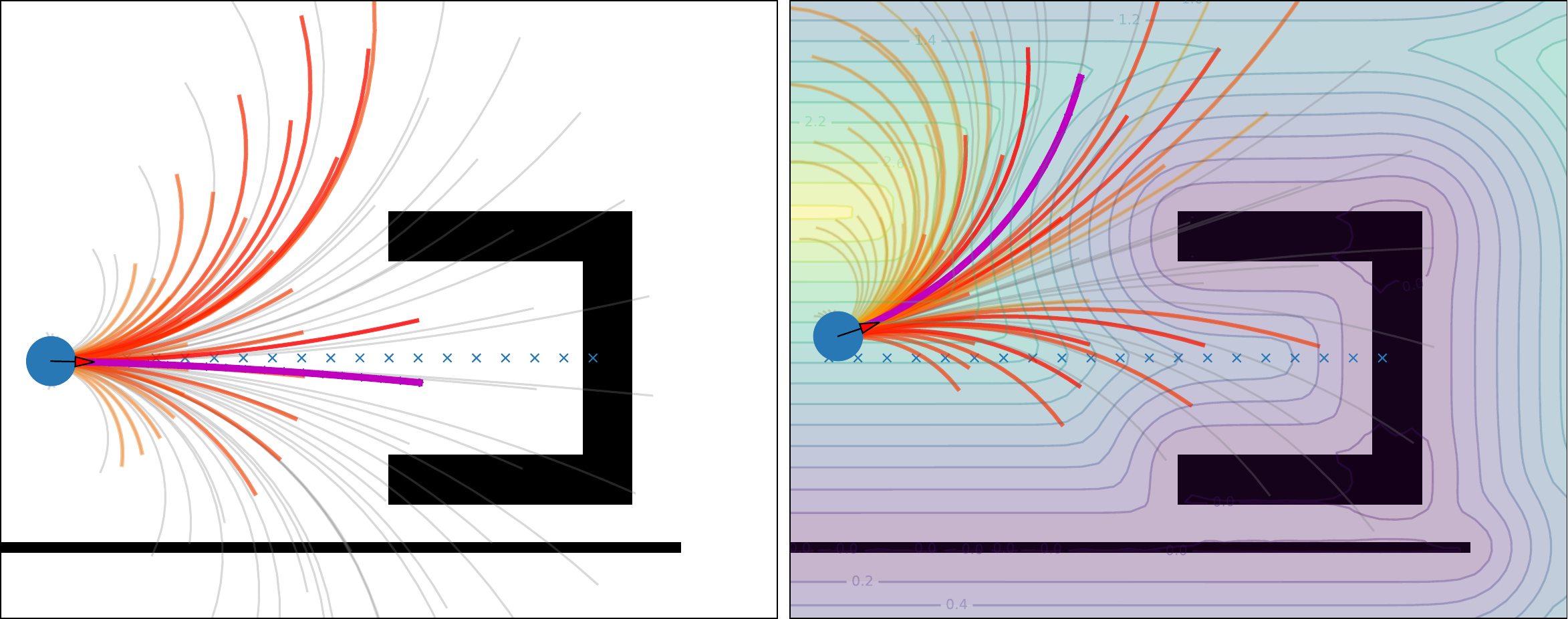}
    \caption{Comparison of trajectory evaluation strategies between the regular DWA (left) and the proposed GF-DWA (right) when encountering the same non-convex obstacle. Black polygons represent static obstacles, while blue circles denote mobile robots, with arrows indicating their movement direction. For trajectory visualization: Gray trajectories are infeasible. Orange trajectories are feasible (with a color gradient: redder hues indicate lower cost, while yellower hues indicate higher cost). Purple trajectories represent the optimal selection.}
    \label{fig:concept}
\end{figure}

Escaping from a non-convex obstacle requires more than just instantaneous distance information; it is essential to determine whether the mobile robot is approaching the obstacle. A straightforward solution is to leverage the gradient of the distance to the obstacle, allowing the robot to take actions that guide it along the gradient direction, thereby avoiding potential collisions.
The Euclidean Signed Distance Field (ESDF) \cite{oleynikova_2016_sdf, Han_2019_esdf} is a widely used method to model the environment, as it provides both distance and gradient information relative to the nearest occupied cell. However, to capture the overall shape of the obstacle, a robot needs to consider the distance and gradient information from all directions, which is crucial for non-convex obstacle avoidance. 
The Gaussian Process Distance Field (GPDF) \cite{le_2023_gpdf, choi_2024_gpdf} offers a probabilistic approach to modeling the environment, providing distance and gradient information from all detected obstacles. This enables a more comprehensive representation of non-convex structures, allowing for more effective motion planning and obstacle avoidance in complex environments.

In this work, we propose the Gradient Field-based Dynamic Window Approach (GF-DWA), a simple yet effective extension of the traditional DWA, incorporating the gradient information from the GPDF in the objective function. 
By leveraging the unified distance and gradient fields from the GPDF, the GF-DWA provides a more accurate and comprehensive representation of the environment, allowing robots to avoid non-convex obstacles effectively. Furthermore, the GF-DWA can be seamlessly extended to the fleet collision avoidance problem, where each robot regards the predicted trajectories of other robots as obstacles and avoids collisions in a distributed manner. The main contributions of this work are:
\begin{itemize}[leftmargin=*]
    \item Introducing a new cost term based on the gradient information from GPDF into DWA, enhancing the robot's ability to avoid non-convex obstacles effectively and efficiently.
    \item Extending GF-DWA to fleet collision avoidance, where predicted trajectories of other robots are modeled using GPDFs in real-time. This allows for a proactive and distributed approach to collision avoidance within multi-robot systems.
\end{itemize}

\section{Preliminaries}
\subsection{Dynamic Window Approach}
As a sampling-based local motion planning and control method, the DWA algorithm \cite{fox_1997_dwa} generates a series of motion plans as candidates by sampling control parameters within a local dynamic window and then selects the one with the lowest cost. To limit the number of candidates, each sampled control action is assumed to stay constant within the window. All candidates are evaluated by a pre-defined objective function, typically comprising collision-avoidance and target-tracking terms. At each time step, the local window shifts according to the current state of the robot, and the best action is adopted to guide the robot. This process loops until the robot reaches the target or halts due to collisions and deadlocks, which is a receding horizon control strategy. At each iteration, the DWA has the following steps:
(i) dynamic window generation, (ii) candidate trajectory generation, and 
 (iii) candidate evaluation.

In order to generate trajectories, a motion model with kinematic constraints is necessary. For the mobile robot navigation task, assuming the discrete motion model is $\bm{x}_{k+1}=f(\bm{x}_{k}, \bm{u}_{k})$, where $\bm{x}_{k}$ is the state and $\bm{u}_{k}$ is the action at time step $k$.The robot state $\bm{x}_{k}=[x,y,\theta]^\top$ consists of the location $(x,y)$ and the heading $\theta$, and the action space $\bm{u}_{k}=[v, \omega]^\top$ contains the linear speed $v$ and the angular velocity $\omega$. The control action is restraint by $(\bm{u}_{\text{max}}, \bm{u}_{\text{min}}, \Delta\bm{u}^+_{\text{max}}, \Delta\bm{u}^-_{\text{max}})$, which defines the bound of the action, $\bm{u}_{\text{max}} \le \bm{u} \le \bm{u}_{\text{min}}$, and its changing rate $\Delta\bm{u}^-_{\text{max}} \le \Delta\bm{u} \le \Delta\bm{u}^+_{\text{max}}$. At time $k$, the dynamic window for control action ranges over the interval $(\bm{u}^-_k, \bm{u}^+_k)$, where
\begin{align}
    \bm{u}^{-}_k &= \max(\bm{u}_{\text{min}}, \bm{u}_{k-1}-\Delta\bm{u}^{-}_{\text{max}}), \\
    \bm{u}^{+}_k &= \min(\bm{u}_{\text{max}}, \bm{u}_{k-1}+\Delta\bm{u}^{+}_{\text{max}}). 
\end{align}
Given the dynamic window horizon $N$, and the sampling number $\bm{\gamma}=[\gamma_v, \gamma_\omega]^\top$ of the action, a set $\mathcal{C}$ of control actions can be generated as
\begin{align*}
    \mathcal{C} = \Bigl\{
        \bm{u}^{(i,j)}_k \, | \,
        &i\in\{0, 1, \ldots, \gamma_v-1\}, \\
        &j\in\{0, 1, \ldots, \gamma_\omega-1\}, \\
        &\bm{u}^{(i,j)}_k=\bm{u}^{-}_k + 
        \begin{bmatrix}
            i/(\gamma_v-1) \\
            j/(\gamma_\omega-1)
        \end{bmatrix}\odot(\bm{u}^{+}_k - \bm{u}^{-}_k)
    \Bigr\},
\end{align*}
where $\bm{a}\odot\bm{b}$ means element-wise product. For each $\bm{u}^{(i,j)}_k\in\mathcal{C}$, a candidate trajectory is then obtained,
\begin{align*}
    \mathcal{T}^{(i,j)}_k = \Bigl\{
        \bm{x}^{(i,j,n)}_k \,| \, 
        &n\in\{0,1,\ldots,N\}, \\
        &\bm{x}^{(i,j,0)}_k = \bm{x}_k, \\
        &\bm{x}^{(i,j,n)}_k |_{n>0}=f(\bm{x}^{(i,j,n-1)}_k,\bm{u}^{(i,j)}_k).
    \Bigr\}
\end{align*}

Although the design of the objective function varies depending on the specific application, two basic cost terms are usually included: the collision-avoidance term and the target-tracking term. The collision-avoidance term $J_{\text{col}}(\cdot)$ quantifies the risk of collision with obstacles or other robots, often defined as the minimum distance to surrounding obstacles. The target tracking term $J_{\text{tar}}(\cdot)$ determines the trajectory’s progress toward the target, commonly measured as the distance from the trajectory endpoint to the final goal. For any trajectory that results in a collision, the collision-avoidance cost is set to positive infinity, rendering it infeasible. Among all feasible trajectories, the one with the lowest total cost is selected for execution.

\begin{figure}[t]
    \centering
    \includegraphics[width=0.99\linewidth]{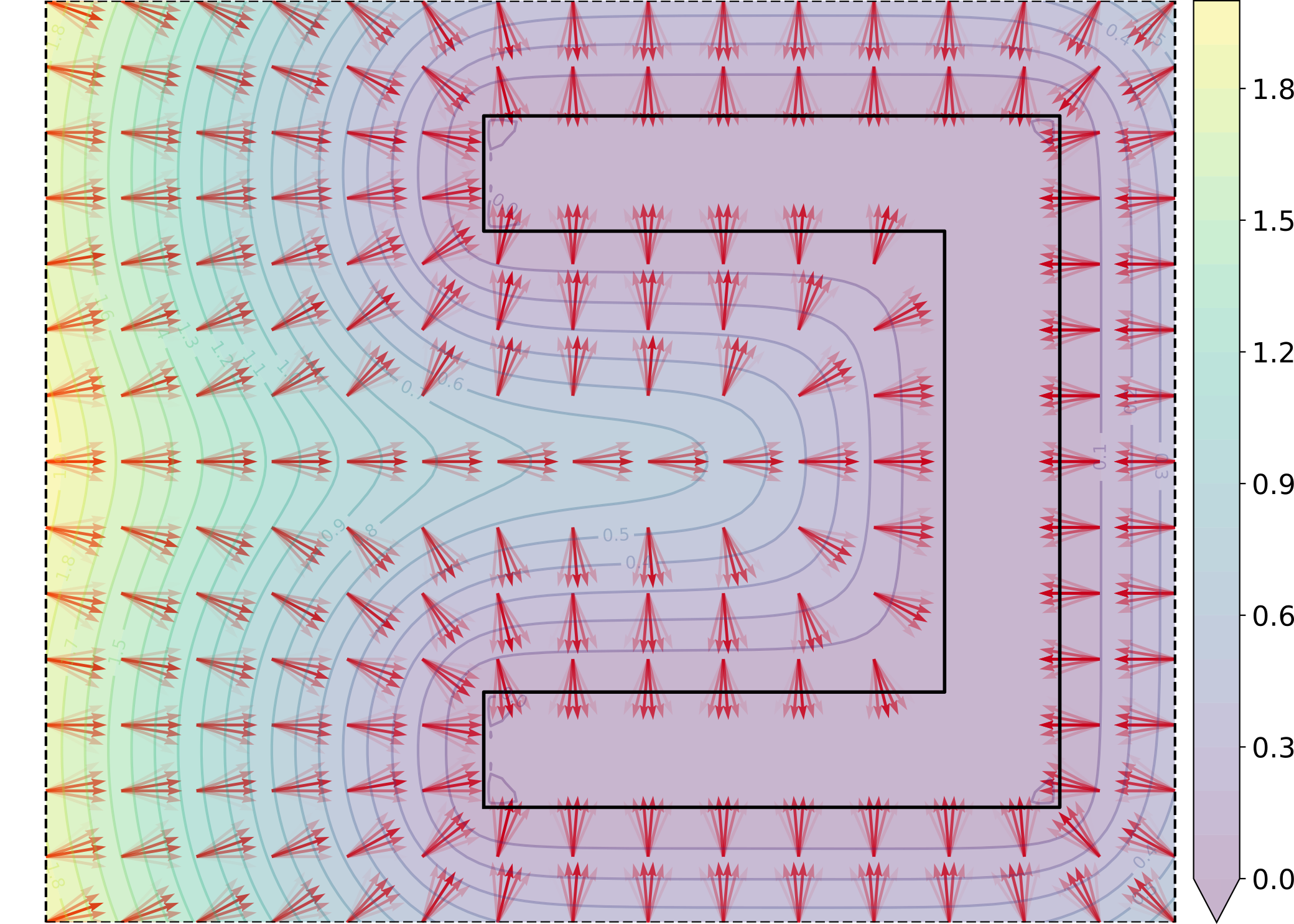}
    \caption{Visualization of the proposed gradient field-based collision avoidance term $J_{\text{col-grad}}$ given $\Delta\theta_{\text{thre}}=2\pi/3$. The black U-shaped polygon represents the obstacle. The contour and color bar illustrate the distance field obtained from the GPDF of the obstacle. Each arrow denotes a robot state, with its tail position indicating the robot’s location and its orientation corresponding to the robot’s heading direction. The arrow color represents the intensity of the cost term, where redder hues indicate a higher cost and vice versa. This cost term effectively prevents the robot from being trapped in local minima.}
    \label{fig:grad_cost}
\end{figure}

\subsection{Gaussian Process Distance Field}
Given a set of points $\mathcal{P}=\{\bm{p}_1, \bm{p}_2, \ldots, \bm{p}_M\}$ representing obstacles, which may be sampled from geometric obstacles or acquired from LiDAR sensory data, each point can be modeled as a Gaussian distribution centered at its measured position, with associated uncertainty. 
Accordingly, a GPDF \cite{choi_2024_gpdf, le_2023_gpdf} can be built to estimate a continuous, differentiable representation of the distance to all detected obstacles at any point in the robot’s workspace.  
Define a latent field $o(\bm{p})$ and the inverse function $f_{\text{inv}}$ mapping to the distance field $f_{\text{dist}}(\bm{p})$,
\begin{align}
    f_{\text{inv}}\left(k_o(\bm{p}, \mathcal{P})\right) &\coloneq ||\bm{p}-\mathcal{P}||, \\
    f_{\text{dist}}(\bm{p}) &= f_{\text{inv}}(o(\bm{p})), \\
    o(\bm{p}) &\sim \mathcal{G}(0, k_o\left(\bm{p}, \mathcal{P})\right),
\end{align}
where $\mathcal{G}$ is Gaussian process and $k_o$ is the covariance kernel function. This latent field can be regarded as a smooth occupancy field. According to the Gaussian process regression,
\begin{align}
    \bar{o}(\bm{p}) &= k_o(\bm{p}, \mathcal{P}) \underbrace{\left(K_o(\mathcal{P}, \mathcal{P})+\sigma_o^2 \mathbf{I}\right)^{-1}\cdot\bm{1}}_{\alpha(\mathcal{P}, \sigma_o)}, \\
    \mathbf{cov}(o(\bm{p})) &= k_o(\bm{p}, \bm{p}) \nonumber\\
    &- k_o(\bm{p}, \mathcal{P})\left(K_o(\mathcal{P}, \mathcal{P})+\sigma_o^2 \mathbf{I}\right)^{-1}k_o(\mathcal{P},\bm{p}),
\end{align}
where $K_o(\mathcal{P}, \mathcal{P})$ is the covariance kernel of the given obstacle points, $\sigma_o$ is the noise covariance, $\mathbf{I}$ is an identity matrix, $\bm{1}$ is a vector of ones (setting $k_o(\mathcal{P}, \mathcal{P})=\bm{1}$), and $\alpha(\cdot)$ is the GP model. The gradient of the distance field can be derived as 
\begin{align}
    f_{\text{grad}}(\bm{p}) &= \nabla_{\bm{p}} f_{\text{dist}}(\bm{p}) \nonumber\\
    &= \frac{\partial f_{\text{inv}}}{\partial o} \cdot \left[
    \nabla_{\bm{p}}k_o(\bm{p},\mathcal{P}) \cdot \alpha(\mathcal{P}, \sigma_o)
    \right].
\end{align}
In the GPDF, the distance field provides direct information about how far a point is from the surrounding obstacles, while the gradient field, obtained by differentiating the distance field, encodes both the direction and rate of change of the distance with respect to the robot's position. 
In the context of collision avoidance, the gradient field effectively acts as a repulsive vector field, which can guide the robot away from obstacles.

\begin{figure}[t]
    \centering
    \includegraphics[width=0.99\linewidth]{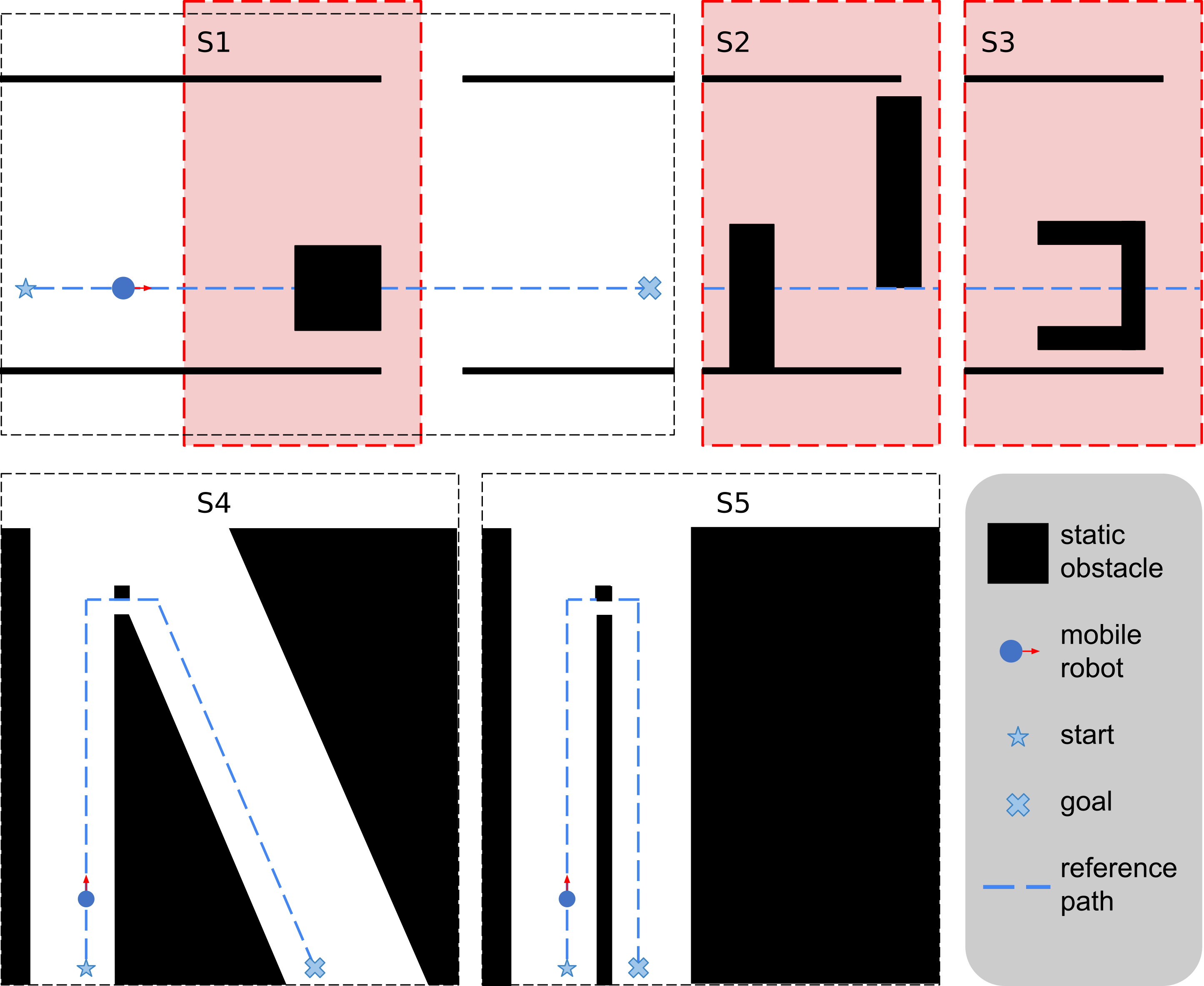}
    \caption{Scenarios with a single robot and static obstacles. The first row demonstrates three scenes (from left to right: S1, S2, and S3) with different static obstacles in the red area. The second row shows two turning scenarios (from left to right: S4 and S5). Note that in turning scenarios, a small rectangle obstacle is put on the reference path, which creates a local minimum.}
    \label{fig:single_scene}
\end{figure}

\section{Gradient-aware Sampling-based Motion Planning}
\subsection{Predictive Obstacle Avoidance based on Gradient Field}
While most DWA-based methods consider only the distance field, this work introduces a new cost term utilizing the gradient information from the GPDF. Considering a candidate $\mathcal{T} = \left\{\bm{x}^{(n)}\right\}^N_{n=0}$, after querying the GP model $\alpha$ and processing the model outputs, a distance set $\{d^{(n)}\}^N_{n=0}$ and a gradient set $\{\bm{g}^{(n)}\}^N_{n=0}$ are obtained for all states in the trajectory. A cost term for being too close to any obstacles can be easily defined based on the distance set (ignoring the current state of the robot),
\begin{equation}
    J_{\text{col-dist}}=1/\min(d^{(1)}, d^{(2)}, \ldots, d^{(N)}).
\end{equation}
This term alone only gives the instant distance information, which is not enough to capture the potential collision risk. For example, a trajectory that is relatively close to an obstacle but not moving toward it is less dangerous than a trajectory that is moving toward the obstacle but is currently further away. To capture this, we introduce a new cost term based on the gradient set,
\begin{equation}
    J_{\text{col-grad}} =
    \begin{cases}
        \sum^N_{n=1}e^{\beta|\Delta \theta|} - 1,\, &|\Delta \theta|\ge\Delta\theta_{\text{thre}} \\
        0, \, &\text{otherwise}
    \end{cases}
\end{equation}
where $\Delta \theta=\theta^{(n)}-\theta^{(n)}_d$ is the difference between the heading angle $\theta^{(n)}$ of the robot and the {\color{revised}normalized} heading direction $\theta^{(n)}_d$ of the gradient at state $\bm{x}^{(n)}$, $\Delta\theta_{\text{thre}}$ is the threshold to include the collision-gradient term, and $\beta$ is a tuning parameter to control changing rate of the cost term. 
The cost term $J_{\text{col-grad}}$ is designed to {\color{revised}provide a smooth, continuous cost increase as heading misalignment grows}.
{\color{revised}The exponential form imposes stronger penalties when the robot moves toward obstacles in its forward direction.}
Although the gradient-based term $J_{\text{col-grad}}$ can capture the potential collision risks, it may lead to overly conservative behavior, causing the robot to avoid obstacles even when not moving toward them. The choice of $\Delta\theta_{\text{thre}}$ plays a crucial role in balancing the trade-off between collision avoidance and regular path following. Consequently, it should be satisfies that $\Delta\theta_{\text{thre}}\in(\pi/2, \pi]$. 
Fig. \ref{fig:grad_cost} visualizes this cost term for a U-shaped non-convex obstacle. The distance field alone creates a local minimum near the dead end, while the gradient-based cost successfully identifies the directions leading to the dead end, preventing the robot from becoming trapped.

The overall collision-avoidance cost term is defined as a combination of the distance-based term and the gradient-based term,
\begin{equation}
    J_{\text{col}} = Q_\text{col-dist} J_{\text{col-dist}} + Q_\text{col-grad} J_{\text{col-grad}},
\end{equation}
where $Q$ variables are the corresponding weights to each term.

\begin{figure*}[t]
    \centering
    \includegraphics[width=0.9\linewidth]{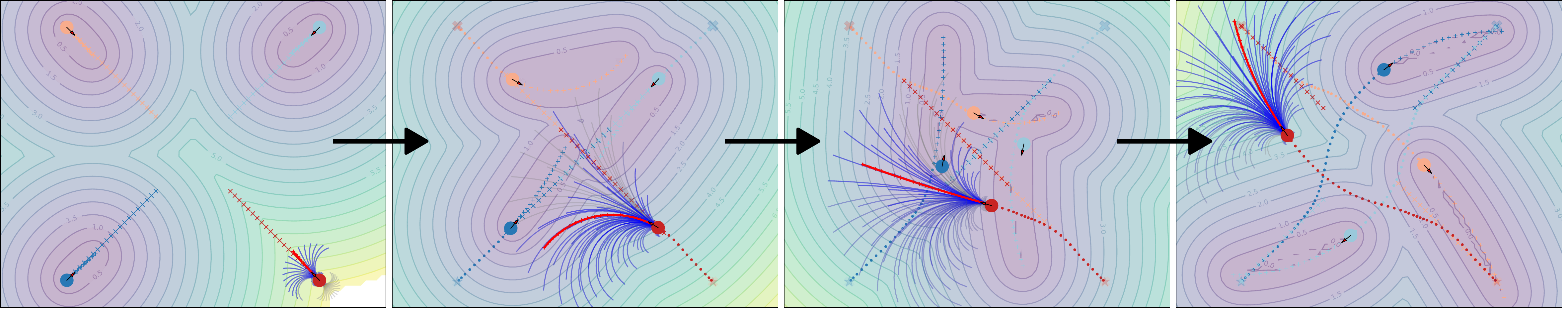}
    \vspace{-3mm}
    \caption{Multi-robot test 1 (using DF-DWA) with four robots intersecting in the middle of the map. This figure shows the candidate trajectories and unified GPDF from the perspective of the robot starting from the bottom right corner. Gray trajectories are infeasible, blue ones are feasible, and red ones are the best in the corresponding time step.}
    \label{fig:multi_1}
\end{figure*}
\begin{figure*}[t]
    \centering
    \includegraphics[width=0.9\linewidth]{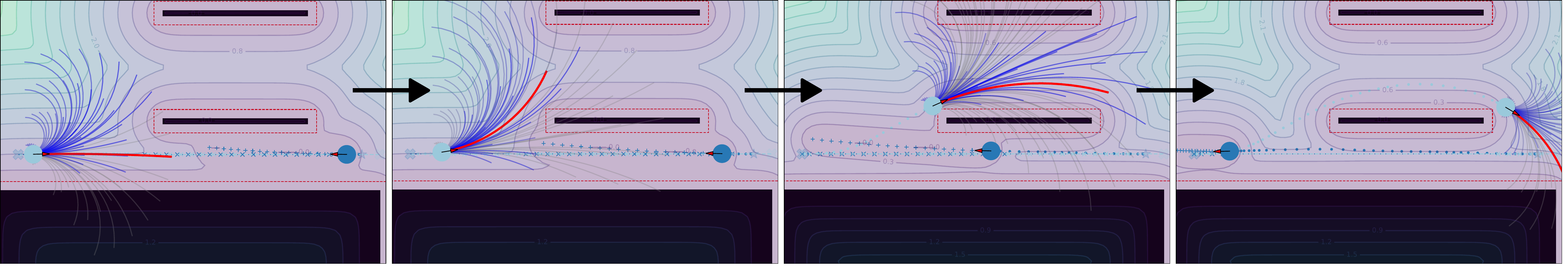}
    \vspace{-3mm}
    \caption{Multi-robot test 2 (using DF-DWA) with two robots with overlapping reference paths meeting in a narrow corridor. This figure shows the candidate trajectories and unified GPDF from the perspective of the robot starting from the left side. Apart from visual elements as in Fig. \ref{fig:single_scene} and Fig. \ref{fig:multi_1}, red dashed lines indicate the inflated boundary of static obstacles by the size of the robot. }
    \label{fig:multi_2}
\end{figure*}

\subsection{Gradient Field-based Dynamic Window Approach}
Apart from the collision-avoidance term, there are also other cost terms that need to be considered in trajectory selection. In this work, it is assumed that a reference path, a target position $\bm{p}_{\text{tar}}=(x_{\text{tar}}, y_{\text{tar}})$, and a reference speed $v_{\text{ref}}$ are given. Based on the current position of the robot, a reference trajectory $\mathcal{T}_{\text{ref},k}$ sampled from the reference path is generated at each time step $k$ (the time step is omitted if there is no ambiguity). To penalize the deviation from the reference trajectory, a cost term $J_{\text{ref}}$ is defined as
\begin{equation}
    J_{\text{ref}} = \frac{1}{N}\sum^N_{n=1}||\bm{x}^{(n)} - \bm{x}_{\text{ref}}^{(n)}||,
\end{equation}
where $\bm{x}_{\text{ref}}^{(n)}$ is the $n$-th state of the reference trajectory. The reference speed deviation is also considered in the cost function as
\begin{equation}
    J_{\text{vel}} = |v - v_{\text{ref}}|.
\end{equation}
For the target-reaching behavior, rather than calculating the Euclidean distance between the robot and the goal position, a cost term $J_{\text{tar}}$ is defined to penalize the deviation from the heading direction to the target position,
\begin{equation}
    J_{\text{tar}} = \left|\tan^{-1}\left(\frac{y_{\text{tar}}-y}{x_{\text{tar}}-x}\right) - \tan^{-1}\left(\frac{y^{(N)}-y}{x^{(N)}-x}\right)\right|,
\end{equation}
where $(x, y)=(x^{(0)}, y^{(0)})$ is the current position of the robot. The target-reaching term is mainly active when the robot is close to the final goal. The overall cost function is defined as a combination of all the cost terms,
\begin{equation}
    J = Q_{\text{col}}J_{\text{col}} + Q_{\text{ref}}J_{\text{ref}} + Q_{\text{vel}}J_{\text{vel}} + Q_{\text{tar}}J_{\text{tar}},
\end{equation}
where $Q$ variables are the corresponding weights. After calculating costs for all candidate trajectories, the trajectory with the minimum cost is selected as the optimal trajectory, i.e., $\mathcal{T}_{\text{opt}} = \mathop{\arg\min}_{\mathcal{T}}J$.

\begin{table}[t]
    \setlength\extrarowheight{3pt}
    \centering
    \caption{Success rates of different methods avoiding different static obstacles. The ``\checkmark'' symbol indicates 100\% success rate, while ``$\times$'' indicates 0\% success rate. The results for DDPG are obtained via 50 runs. The other results are deterministic.}
    \begin{tabular}{|l|*{4}{c|}}\hline
        \backslashbox{Scene}{Method}
        &\makebox[3em]{MPC}&\makebox[3em]{DDPG}&\makebox[3em]{DWA}
        &\makebox[3em]{\textbf{GF-DWA}}\\
        \hline
        S1 - Rectangle obstacle & \checkmark & \checkmark & \checkmark & \checkmark \\\hline
        S2 - Double obstacles   & $\times$   & $\times$   & \checkmark & \checkmark \\\hline
        S3 - U-shape obstacle   & $\times$   & \checkmark & $\times$   & \checkmark \\\hline
        S4 - Sharp turn         & $\times$   & 96\%       & \checkmark & \checkmark \\\hline
        S5 - U-turn             & $\times$   & 74\%       & \checkmark & \checkmark \\\hline
    \end{tabular}
    \label{tab:single_result}
\end{table}

\subsection{Predictive Fleet Collision Avoidance}
In the presence of multiple robots, the collision avoidance problem becomes more complex due to two main reasons. 
First, each robot must actively avoid collisions with other robots, all of which are moving simultaneously. 
Second, the dynamic interaction between other robots and static obstacles can create non-convex obstacles with continuously changing shapes. 
In \cite{missura_2019_preddwa}, it was proposed to consider the predicted trajectories of other robots to avoid collisions, which can also be seamlessly integrated into our GF-DWA framework. 
At each time step, a mobile robot models the predicted trajectories of other robots from the previous time step using GPDFs. These GPDFs are then merged with environmental obstacles to construct a unified distance field and gradient field. 
The GPDFs generated from predicted robot trajectories are continuously updated at each time step or whenever the robots' states change. In this way, fleet collision avoidance is naturally integrated into the GF-DWA framework, allowing for a distributed and scalable solution to multi-robot navigation in dynamic environments.

\section{Implementation}
In this work, the nonholonomic unicycle model is used as the motion model:
\begin{equation}
    \begin{bmatrix}
        x_{k+1} \\
        y_{k+1} \\
        \theta_{k+1}
    \end{bmatrix} =
    \begin{bmatrix}
        x_k + v_k\cos(\theta_k)\Delta t \\
        y_k + v_k\sin(\theta_k)\Delta t \\
        \theta_k + \omega_k\Delta t
    \end{bmatrix},
\end{equation}
where $\Delta t$ is the sampling time and is set to be 0.2 seconds. The covariance kernel function of the GPDF is the Matérn $\nu=1/2$ kernel with $\sigma=1$:
\begin{equation}
    k_o(d) = \sigma^2\exp\left(-\frac{d}{L}\right),
\end{equation}
where $d$ is the distance between two points and $L=0.2$ is a positive parameter controlling the interpolation degree. The boundary points of obstacles are sampled from the obstacle edges at a resolution of 0.1 meters.
For the DWA, the horizon $N$ is 20, the threshold to include the collision-gradient term is set as $\Delta\theta_{\text{thre}}=2\pi/3$, and the tuning parameter is set as $\beta=2$. {\color{revised}The choice of $\Delta\theta_{\text{thre}}$ creates a strong repulsive force when heading toward obstacles, and minimizes impact when moving parallel or away from obstacles.} In addition, to avoid over-conservative behavior, the collision-avoidance cost term is inactive unless any point on the candidate is within a one-meter range of any obstacle. To generate the candidate control set, the speed resolution is 0.3 meters per second and the angular velocity resolution is 0.08 radians per second (about 4.6 degrees per second). The maximal total number of candidate trajectories is 84.
{\color{revised}For the GPDF, after ahead-of-time compilation, on an Intel i7-9750H CPU, the inference time is around 0.008 seconds in every time step, which achieves highly real-time performance. Note that this can be further improved using GPUs.}

\section{Evaluation}
In the evaluation\footnote{The demo video is available at \url{https://youtu.be/Et4C5oGHlF0}}, the proposed \textbf{GF-DWA} is compared with the regular DWA, optimization-based MPC, and learning-based Deep Deterministic Policy Gradient (DDPG) algorithm \cite{lillicrap_2016_ddpg} in different scenarios. The regular DWA shares the same objective function as GF-DWA except for the collision-avoidance term, which is only based on the minimal distance to any obstacles. The implementation of MPC and DDPG is according to \cite{Ceder_2024_ddpg}. For the single-robot cases, the success rates of selected methods are compared. A successful run means the robot can reach the final goal without any collisions. For the multi-robot cases, besides the success rate, the number of time steps needed to finish the case is also compared. Note that the proposed GF-DWA is sampling-based, with no need to solve an optimization problem at each time step or pre-train any models.

\subsection{Static Obstacle Avoidance}
As demonstrated in Fig. \ref{fig:single_scene}, five test scenarios (S1 to S5) are set for the single-robot evaluation. The result of success rates is shown in TABLE \ref{tab:single_result}. 
It is noteworthy that MPC fails in almost all cases due to the presence of local minima, which arise when encountering large or non-convex obstacles. The learning-based DDPG algorithm successfully avoids the non-convex U-shape obstacle but lacks stability when navigating through environments with closely spaced obstacles. From the evaluation, the proposed GF-DWA is the only method that guarantees collision-free navigation in all cases, showing its stability and adaptability in complex environments.

\begin{figure}[t]
    \centering
    \includegraphics[width=0.99\linewidth]{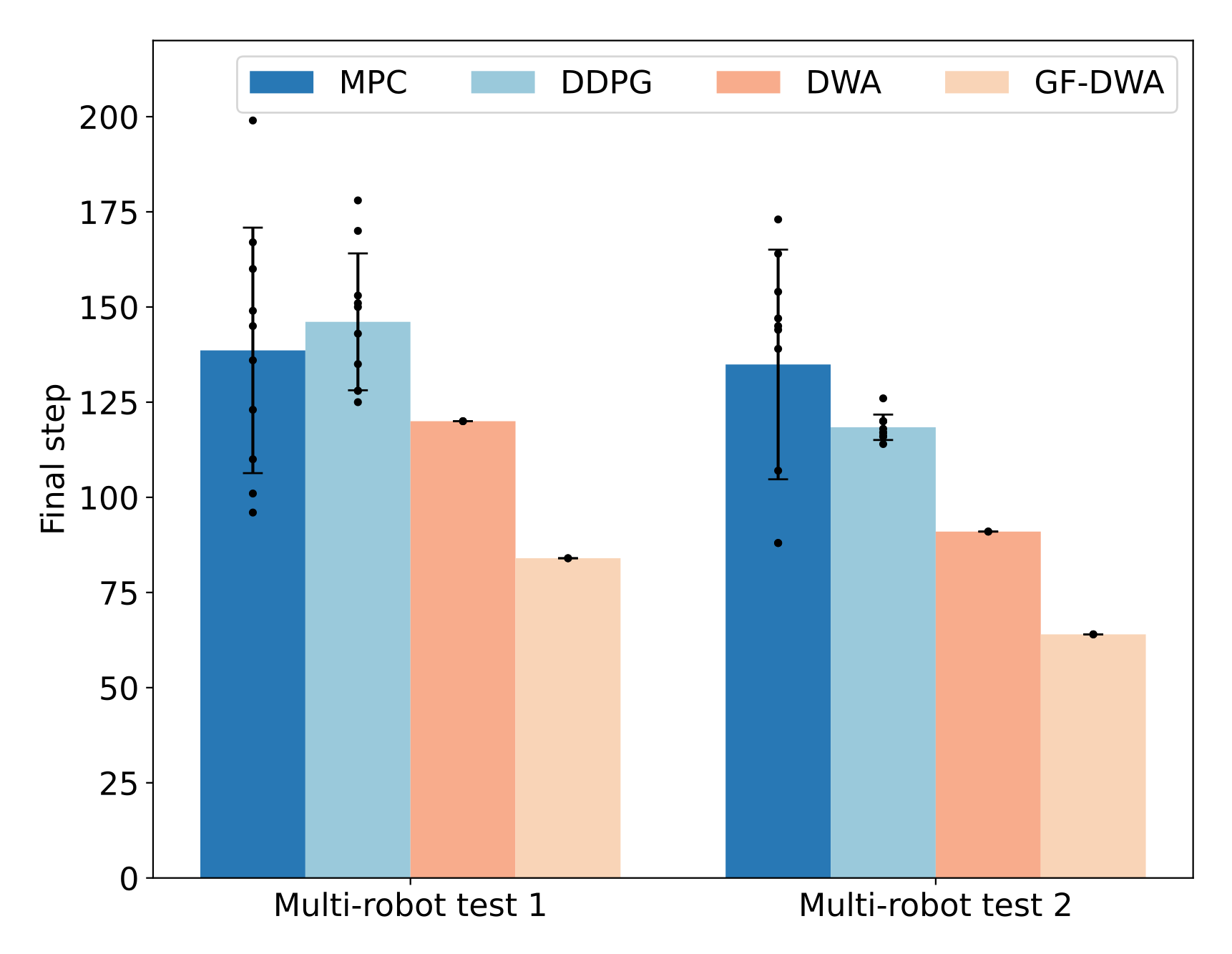}
    \vspace{-5mm}
    \caption{Evaluation of the multi-agent cases. Each bar shows the average time step to reach the goals. Black bars indicate the standard deviations of instances in the category. Black dots are individual trials. The timeout limit is 200 steps, which means MPC has failures in the first case.}
    \label{fig:multi_eva}
\end{figure}

{\color{revised}
\subsection{Gradient information}
To further investigate the effectiveness of the gradient information from GPDFs, additional tests were conducted in S3 with a U-shape obstacle. Variants of DWA were tested using distance-only data from GPDF, gradient data from ESDF, and gradient data from GPDF. Results show that only the GPDF-based gradient enables successful navigation, as ESDF gradients reflect only the nearest obstacle point and fail to capture the global structure of non-convex obstacles.
}

\subsection{Multi-robot collision avoidance}
To evaluate the proposed GF-DWA in multi-robot environments, two scenarios are designed. As shown in Fig. \ref{fig:multi_1}, the first scenario includes four robots starting from the corners of an empty map and moving to the opposite corners, which leads to an intersection at the center of the map. At this intersection, each robot perceives the other three robots as a dynamically changing non-convex obstacle, making collision avoidance more challenging. 
The second scenario, as illustrated in Fig. \ref{fig:multi_2}, involves two mobile robots moving in opposite directions through a narrow corridor. While the corridor has the capacity of two robots side by side, navigating through it simultaneously poses a collision risk. As an alternative, a wider parallel path is available, requiring the robots to decide whether to pass through the corridor or take the wider path to ensure collision-free navigation.

Unlike the single-robot static obstacle avoidance tests, all compared methods can handle the multi-robot tests but with varying levels of efficiency and stability. The detailed finishing time step of each trial is reported in Fig. \ref{fig:multi_eva}. While DWA and GF-DWA generate deterministic control actions, MPC exhibits instability since it requires numerical solvers in practice. 
In multi-robot test 1, {\color{revised}for the regular DWA}, the intersection distance between different robots is very small, posing a significant collision risk in practical deployment. On the contrary, GF-DWA produces predictive collision avoidance behavior as shown in Fig. \ref{fig:multi_1}, ensuring a safer and more stable navigation strategy. 
Overall, GF-DWA outperforms other methods by achieving a better balance between stability, safety, and efficiency, making it a more robust solution for multi-robot collision avoidance in complex environments.

\section{Conclusion}
In this work, we propose a Gradient Field-Based Dynamic Window Approach (GF-DWA) for mobile robot navigation in complex environments and extend it to fleet collision avoidance with predictive behavior in a distributed manner. 
While the underlying principle is straightforward, the integration of unified gradient information from the Gaussian process distance field makes the approach safe, stable, effective, and flexible, particularly in non-convex obstacle avoidance. 
Extensive evaluations, including both single-robot and multi-robot settings, demonstrate that our approach outperforms other popular obstacle avoidance approaches, including optimization-based and learning-based methods, in terms of efficiency, stability, and collision-free navigation.

Future work would focus on three aspects. The first one is to consider dynamic obstacle (such as humans) avoidance in the method. Unlike mobile robots, the future states of dynamic obstacles are typically unknown, posing a significant challenge for predictive navigation.
Another future work is to explore the combination of GF-DWA with other planning and control methods, such as model predictive control, for leveraging the advantages of both sampling-based and optimization-based approaches for improved motion planning and collision avoidance.
The last one is to conduct real-world evaluations of the proposed approach in a physical multi-robot fleet scenario to assess its practical applicability, robustness, and scalability in dynamic and unstructured environments.



\bibliographystyle{IEEEtran}
\bibliography{IEEEabrv, main}

\end{document}